\documentclass[10pt,twocolumn,letterpaper]{article}

\usepackage[pagenumbers]{cvpr} %

\usepackage{multirow}

\newcommand{\second}[1]{\colorbox[HTML]{FFDA8F}{#1}}
\newcommand{\first}[1]{\colorbox[HTML]{FE996B}{#1}}

\usepackage{bm}
\usepackage{booktabs}

\usepackage{stfloats}

\definecolor{ao}{rgb}{0.0, 0.0, 1.0}
\definecolor{airforceblue}{rgb}{0.36, 0.54, 0.66}
\definecolor{ceruleanblue}{rgb}{0.16, 0.32, 0.75}
\definecolor{cerulean}{rgb}{0.0, 0.48, 0.65}
\definecolor{celestialblue}{rgb}{0.29, 0.59, 0.82}
\definecolor{azure(colorwheel)}{rgb}{0.0, 0.5, 1.0}
\definecolor{babyblue}{rgb}{0.54, 0.81, 0.94}
\definecolor{babyblueeyes}{rgb}{0.63, 0.79, 0.95}
\definecolor{ballblue}{rgb}{0.13, 0.67, 0.8}

\definecolor{asparagus}{rgb}{0.53, 0.66, 0.42}
\definecolor{ao(english)}{rgb}{0.0, 0.5, 0.0}
\definecolor{applegreen}{rgb}{0.55, 0.71, 0.0}
\definecolor{armygreen}{rgb}{0.29, 0.33, 0.13}
\definecolor{gray-asparagus}{rgb}{0.27, 0.35, 0.27}
\definecolor{green(ryb)}{rgb}{0.4, 0.69, 0.2}

\definecolor{amethyst}{rgb}{0.6, 0.4, 0.8}
\definecolor{antiquefuchsia}{rgb}{0.57, 0.36, 0.51}
\definecolor{blue-violet}{rgb}{0.54, 0.17, 0.89}
\definecolor{brightlavender}{rgb}{0.75, 0.58, 0.89}
\definecolor{brightube}{rgb}{0.82, 0.62, 0.91}
\definecolor{brilliantlavender}{rgb}{0.96, 0.73, 1.0}

\definecolor{amber}{rgb}{1.0, 0.75, 0.0}
\definecolor{amber(sae/ece)}{rgb}{1.0, 0.49, 0.0}
\definecolor{atomictangerine}{rgb}{1.0, 0.6, 0.4}
\definecolor{burntorange}{rgb}{0.8, 0.33, 0.0}
\definecolor{burntsienna}{rgb}{0.91, 0.45, 0.32}
\definecolor{cadmiumorange}{rgb}{0.93, 0.53, 0.18}
\definecolor{carrotorange}{rgb}{0.93, 0.57, 0.13}
\definecolor{chocolate(web)}{rgb}{0.82, 0.41, 0.12}
\definecolor{chromeyellow}{rgb}{1.0, 0.65, 0.0}
\definecolor{darkorange}{rgb}{1.0, 0.55, 0.0}
\definecolor{darktangerine}{rgb}{1.0, 0.66, 0.07}
\definecolor{deepcarrotorange}{rgb}{0.91, 0.41, 0.17}
\definecolor{deepsaffron}{rgb}{1.0, 0.6, 0.2}
\definecolor{fulvous}{rgb}{0.86, 0.52, 0.0}

\newcommand{\methodname}{3DiMo}
\newcommand{\myparagraph}[1]{\noindent\textbf{#1}}

\usepackage{multirow}
\usepackage[table,xcdraw]{xcolor}

\usepackage{marvosym}

\definecolor{cvprblue}{rgb}{0.21,0.49,0.74}
\usepackage[pagebackref,breaklinks,colorlinks,allcolors=cvprblue]{hyperref}

\def\paperID{***} %
\def\confName{CVPR}
\def\confYear{2026}

\title{
\vspace{-25pt}
3D-Aware Implicit Motion Control for View-Adaptive Human Video Generation
\vspace{-15pt}
}

\author{
Zhixue Fang$^{1,*}$ \quad Xu He$^{2,*}$ \quad Songlin Tang$^{1,*}$ \quad Haoxian Zhang$^{1,\dag,\textrm{\Letter}}$ \\
Qingfeng Li$^{3}$ \quad Xiaoqiang Liu$^{1}$ \quad Pengfei Wan$^{1}$ \quad Kun Gai$^{1}$
\vspace{1mm} \\
$^{1}$Kling Team, Kuaishou Technology \quad
$^{2}$Tsinghua University \quad
$^{3}$CASIA \\
\vspace{1mm}
{$^{*}$Equal contribution \quad $^{\dag}$Project leader \quad $^{\textrm{\Letter}}$Corresponding author} \\
{\url{https://hjrphoebus.github.io/3DiMo}}
}

\begin{document}

\twocolumn[{
\maketitle
\begin{center}
    \vspace{-.3in}
    \captionsetup{type=figure}
    \includegraphics[width=1.0\linewidth]{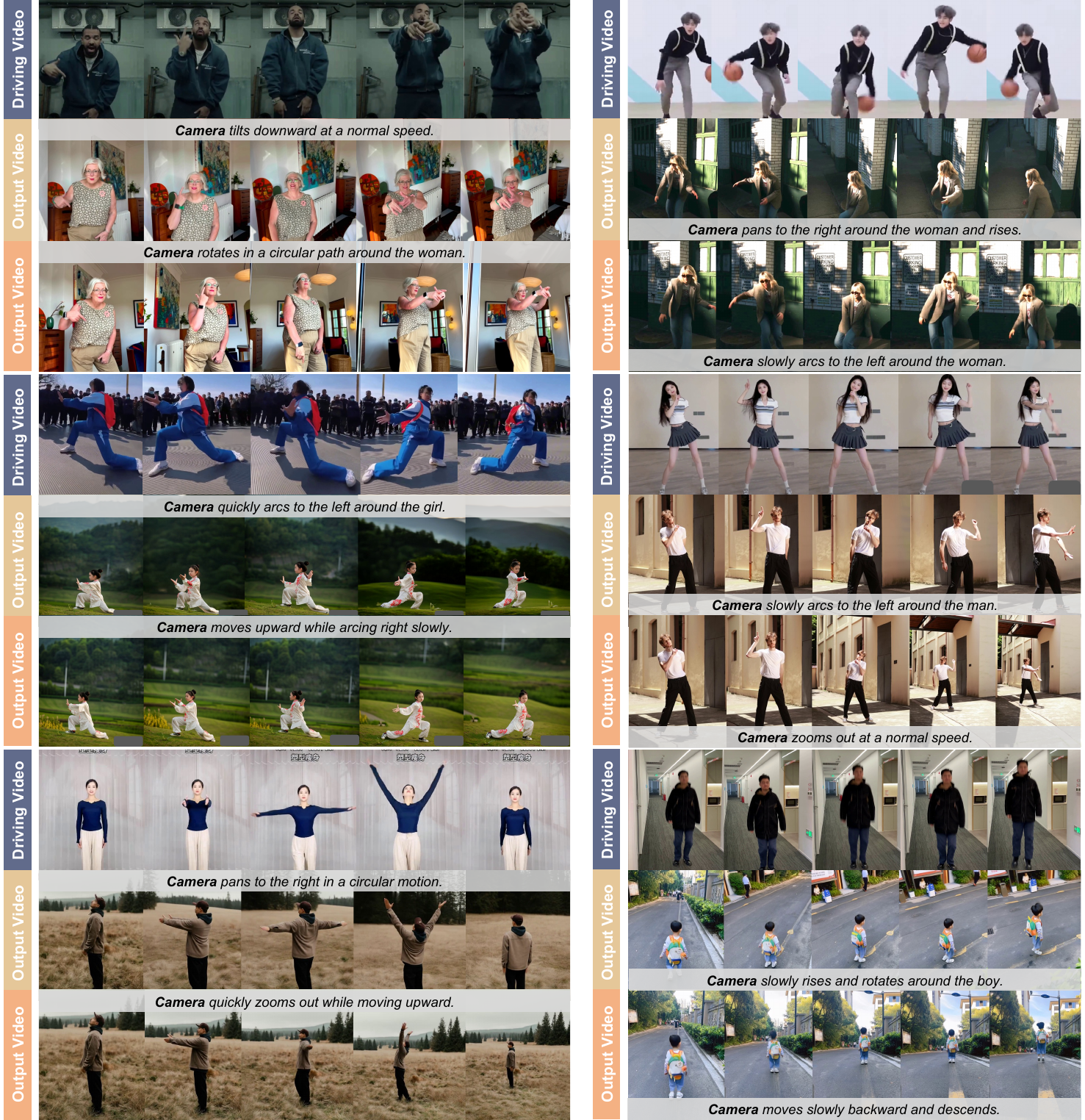}
    \vspace{-.2in}
    \captionof{figure}{
    \textbf{\methodname{}} can faithfully reproduce the 3D spatial motion from a 2D driving video, supporting flexible text-guided camera control. 
    }
    \label{fig:teaser}
\end{center}
}]

\begin{abstract}
Existing methods for human motion control in video generation typically rely on either 2D poses or explicit 3D parametric models (e.g., SMPL) as control signals.
However, 2D poses rigidly bind motion to the driving viewpoint, precluding novel-view synthesis. Explicit 3D models, though structurally informative, suffer from inherent inaccuracies (e.g., depth ambiguity and inaccurate dynamics) which, when used as a strong constraint, override the powerful intrinsic 3D awareness of large-scale video generators.
In this work, we revisit motion control from a 3D-aware perspective, advocating for an implicit, view-agnostic motion representation that naturally aligns with the generator's spatial priors rather than depending on externally reconstructed constraints.
We introduce 3DiMo, which jointly trains a motion encoder with a pretrained video generator to distill driving frames into compact, view-agnostic motion tokens, injected semantically via cross-attention.
To foster 3D awareness, we train with view-rich supervision---single-view, multi-view, and moving-camera videos---forcing motion consistency across diverse viewpoints. 
Additionally, we use auxiliary geometric supervision that leverages SMPL only for early initialization and is annealed to zero, enabling the model to transition from external 3D guidance to learning genuine 3D spatial motion understanding from the data and the generator's priors.
Experiments confirm that 3DiMo faithfully reproduces driving motions with flexible, text-driven camera control, significantly surpassing existing methods in both motion fidelity and visual quality.
\end{abstract}

\section{Introduction}
\label{sec:intro}
Recent advances show that large-scale video generation models possess strong 3D spatial awareness and motion reasoning~\cite{tong2025thinking,wiedemer2025video}, enabling text-guided novel-view synthesis and human reposing with consistent 3D geometry~\cite{wan2025wan,kong2024hunyuanvideo,li2024sora}.
Meanwhile, controllable video generation has emerged as a research focus, with human motion control being one of its central challenges, which aims to animate a reference image according to motion cues from a driving video.
Existing approaches typically extract 2D-rendered pose images from driving frames~\cite{zhang2024mimicmotion,zhu2024champ,cao2025uni3c,ding2025mtvcrafter,he2024co} and inject them via pixel-aligned conditioning~\cite{zhang2023adding,hu2024animate}.
However, such 2D conditioning rigidly binds motion to the driving viewpoint, preventing the model from reasoning about motion in its inherently 3D nature.
As a result, generated videos collapse to the 2D projection of the driving view, losing viewpoint flexibility and limiting applications such as novel-view human synthesis or cinematic camera motion.

In this work, we revisit the task of human motion control from a 3D-aware perspective. 
Our goal is to enable the model to reproduce the underlying 3D motion implied in 2D driving frames, while maintaining independent, text-guided camera control during generation.
To achieve similar goals, recent works attempt to explicitly separate motion and camera control via 3D reconstruction---recovering SMPL(-X)~\cite{SMPL:2015,SMPL-X:2019} sequences from driving videos and conditioning generation through mesh rendering~\cite{cao2025uni3c} or projected keypoints~\cite{ding2025mtvcrafter} under predefined camera trajectories.
This line of work indeed moves beyond 2D constraints by recognizing that human motion inherently occurs in 3D space.
However, it still faces a fundamental limitation: the motion representation is fully determined by externally estimated parametric reconstructions such as SMPL. These estimates, while structurally stable, suffer from depth ambiguities~\cite{he2025magicman} (e.g., forward tilting, incorrect inter-limb contact, or distorted Z-axis motion), and their inaccurate reconstruction of natural motion dynamics further limits expressiveness.
When such biased 3D signals are injected into the generator---especially through rigid projection-based 2D alignment---they impose strong geometric constraints that override native 3D priors of large-scale video models, ultimately limiting the generator's ability to produce spatially coherent and physically plausible motion.

To address these limitations, we propose a new paradigm of \emph{implicit 3D reasoning for motion control} that leverages the video generator's intrinsic spatial and motion understanding.
We argue two key principles.
First, we advocate for end-to-end learning of a motion encoder jointly with the generator, extracting implicit 3D motion representations directly from 2D frames while naturally aligning with the model's spatial priors, which requires encoder designs that encourage view-agnostic motion discovery with semantic conditioning rather than rigid projection.
Second, effective 3D awareness demands supervision beyond conventional same-view reconstruction, which merely learns 2D projection patterns.
Instead, view-rich data across diverse viewpoints and camera trajectories forces the extraction of the essential 3D spatial motion.
Unlike works such as Uni3C~\cite{cao2025uni3c}, which emphasize precise camera trajectory control in animation scenarios, our work focuses on modeling and reproducing 3D motion from 2D observations.
In our framework, camera control is not an explicit objective but a natural byproduct of the model's learned 3D awareness.
We therefore leverage the generator's native text-driven camera control---rather than predefined camera parameters---as it both aligns with the model's intrinsic spatial understanding and serves as evidence of whether genuine 3D awareness has emerged in the learned motion representations.

Building on this paradigm, we present \textbf{\methodname{}}, an end-to-end framework to learn \textbf{3D}-aware \textbf{i}mplicit \textbf{mo}tion control for view-adaptive human video generation under view-rich supervision.
Specifically, we design a Transformer-based motion encoder that distills 2D driving frames into compact 1D tokens, intentionally discarding spatial layout to encourage viewpoint-agnostic, semantic motion abstraction.
The encoder is jointly optimized with a pretrained video generator to align with the generator's generative capability.
The resulting motion tokens are injected through cross-attention, enabling flexible semantic conditioning in place of rigid projection-based alignment.
To achieve genuine 3D awareness, we collect and train on a comprehensive view-rich dataset spanning single-view, multi-view, and moving-camera videos.
Each clip produces motion cues that condition the generator, which is supervised to either reconstruct the same video or reproduce the motion from alternative viewpoints or camera trajectories guided by text prompts.
This dual-objective training encourages the emergence of expressive, 3D-aware motion representations.

To accelerate spatial understanding during early training, we further introduce lightweight auxiliary decoders that provide geometric supervision by aligning motion features with parametric 3D reconstruction results (\ie, SMPL and MANO~\cite{MANO:SIGGRAPHASIA:2017}).
Although these external estimates are imperfect, they supply 3D human priors that offer a reliable initialization.
As training progresses, the auxiliary loss is gradually annealed to zero, allowing the model to shift from externally guided geometry to the generator's inherent 3D priors and the richness of view-rich data---ultimately enabling expressive and truly 3D-aware motion representations.

Our contributions can be summarized as:
\begin{itemize}
    \item \textbf{3D-aware motion control.}
    We reformulate human motion control for video generation as a 3D-aware task that recovers underlying 3D motion from 2D frames while naturally supporting flexible text-driven camera control.
    
    \item \textbf{End-to-end implicit motion framework.}
    We propose \textbf{\methodname{}}, an end-to-end framework that jointly learns a view-agnostic implicit motion encoder with a powerful pretrained DiT-based video generator. This design encourages motion representations that align with the generator's intrinsic 3D spatial priors and enables semantically rich motion conditioning via cross-attention.
    
    \item \textbf{View-rich supervision for 3D learning.} 
    We collect a large-scale human motion dataset spanning single-view, multi-view, and moving-camera videos, enabling viewpoint-agnostic 3D motion learning aligned with the generator’s inherent 3D reasoning. The collected subset will be released to support future research.
\end{itemize}
Through extensive experiments, we demonstrate that \methodname{} faithfully reproduces driving motions while preserving 3D consistency across varying viewpoints, validating that the learned motion representations are both expressive and 3D-aware, effectively conditioning the DiT-based video model to generate high-fidelity human motion videos.

\section{Related Work}
\myparagraph{Diffusion-Based Video Generation.}
Diffusion models have achieved remarkable success in high-fidelity image and video synthesis.
Latent Diffusion Models (LDMs)~\cite{rombach2022sd,blattmann2023align} improve efficiency by operating in compressed latent spaces, while DiT-based architectures~\cite{peebles2023scalable} further enhance scalability and spatiotemporal consistency for video generation.
Recent advances~\cite{li2024sora,kuaishou2024kling,midjourney2024,wan2025wan} show that large-scale pretrained video diffusion models exhibit strong awareness and reasoning capabilities over both dynamics and 3D space~\cite{wiedemer2025video,tong2025thinking}.
In this work, we focus on learning an implicit 3D-aware motion representation that aligns with a pretrained video generator's spatial and motion priors---thereby eliciting their intrinsic 3D understanding and enabling high-quality, spatially consistent human animation.

\myparagraph{Motion Control for Human Image Animation.}
Human image animation aims to animate a reference image according to motion cues from a driving video, as pioneered by early works such as FOMM~\cite{fomm} and MRAA~\cite{mraa}.
Recent diffusion-based approaches~\cite{hu2024animate,zhang2024mimicmotion} achieve impressive quality by injecting explicit motion signals (e.g., 2D poses or DensePose), but their 2D formulations inherently lose spatial information, causing depth ambiguities.
To overcome this, 3D-based methods~\cite{zhu2024champ,luo2025dreamactor,ding2025mtvcrafter,cao2025uni3c,he2025magicman} introduce SMPL~\cite{SMPL:2015} or SMPL-X~\cite{SMPL-X:2019} models as control conditions—typically rendered or projected into 2D space, or mapped as camera-space joint trajectories for explicit 3D control.
However, such approaches rely heavily on externally reconstructed representations, which, though structurally stable, lack the expressiveness and 3D reasoning priors present in large-scale pretrained video generators.
In contrast, we learn an implicit, end-to-end motion representation aligned with these priors to achieve expressive and 3D-aware motion modeling.
While X-Nemo~\cite{zhao2025xnemo} and X-UniMotion~\cite{song2025xuni} explore implicit motion representations, they remain limited to 2D spatial patterns and cannot generalize to true 3D motion or camera control—challenges that our work directly addresses.

\begin{figure*}[t]
\centering
\includegraphics[width=\textwidth]{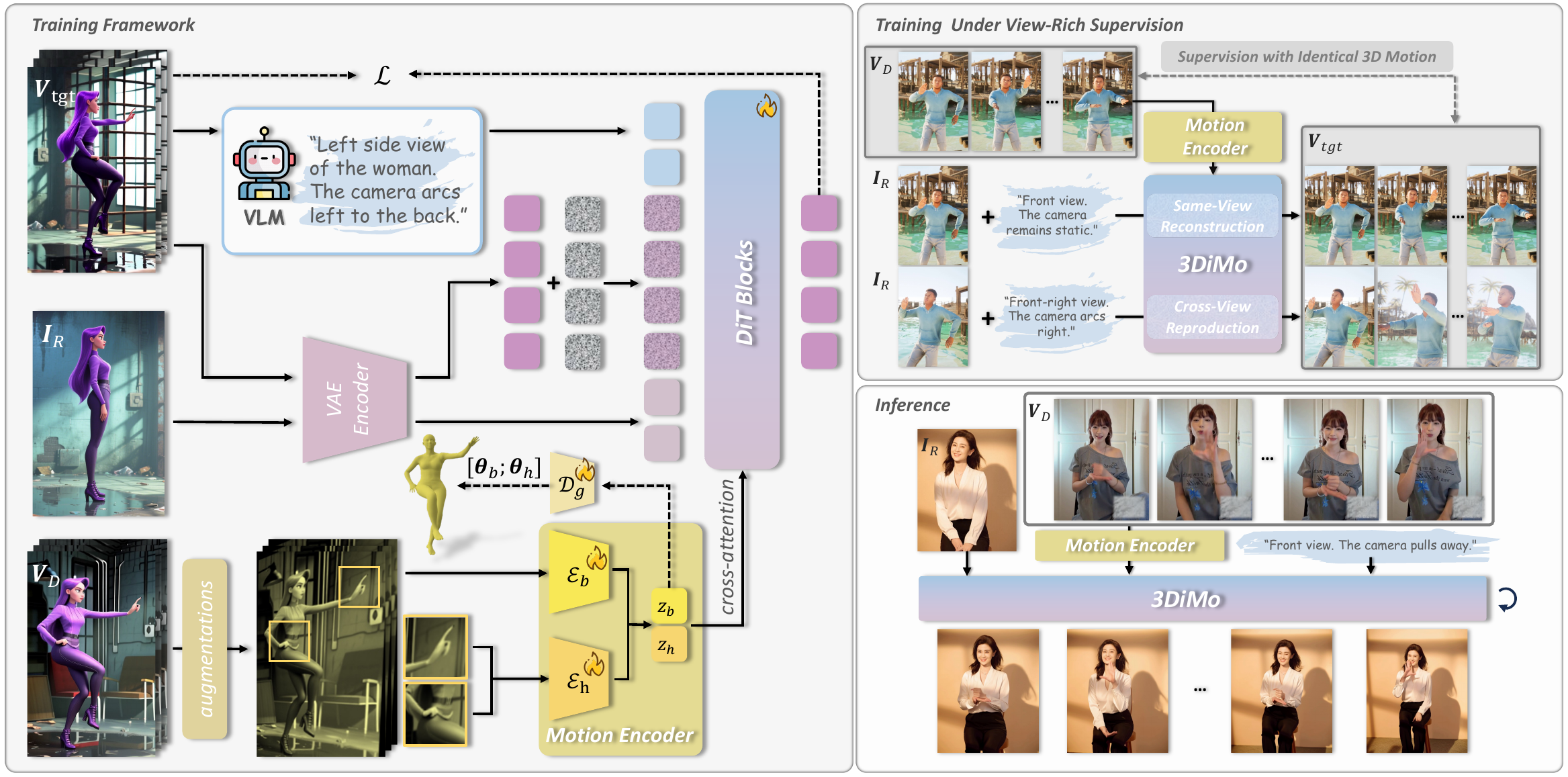}
\caption{
\textbf{Overview of \methodname{}.}
Our framework consists of end-to-end trained motion encoders---$\mathcal{E}_b$ for the body and $\mathcal{E}_h$ for hands---and an DiT-based video generator.
Given a reference frame $\bm{I}_R$ and a driving video $\bm{V}_D$, driving frames are first augmented with random perspective transformations before being encoded by the motion encoder to extract view-agnostic motion representations.
These resulting features are then injected into the generator through cross-attention, enabling the model to synthesize a target sequence $\bm{V}_{\mathrm{tgt}}$ that reenacts the same underlying 3D motion while preserving flexible text-driven camera control.
To facilitate 3D-aware learning, we introduce early-stage auxiliary geometric supervision by regressing the encoded motion to external parametric reconstruction results $\theta_b$ and $\theta_h$.
During training, view-rich data is used to jointly supervise same-view reconstruction and cross-view motion reproduction, driving the emergence of expressive and 3D-aware motion representations.
At inference, motion tokens extracted directly from 2D driving frames provide rich 3D spatial cues that can animate any reference character, supporting high-fidelity and view-adaptive motion-controlled video generation.
}
\label{fig:pipeline}
\end{figure*}

\section{Our Approach}
Given a reference image $\bm{I}_R$ of a subject and a driving video $\bm{V}_D=\{\bm{I}_D^t\}_{t=0}^T$ providing motion cues, our proposed 3D-aware motion control aims to transfer the driving video's motion---which inherently exists in 3D space---to the reference subject, while preserving flexible, text-guided camera control.
This task is highly challenging as human motion and camera trajectories are entangled within the 2D projections of driving frames, obscuring the underlying 3D motion essence that truly exist in physical space.

To achieve this goal, we leverage a pretrained DiT-based video generation model with rich 3D spatial and motion priors as our backbone, which generates videos from a reference image guided by text prompts (\cref{sec:method_preliminary}).
The core of our framework lies in an implicit motion encoder jointly optimized with the pretrained video generator, which distills view-agnostic motion tokens from 2D driving frames and injects them via cross-attention for semantical motion control compatible with text-driven camera manipulation (\cref{sec:method_pipelie}).
To endow the learned motion representation with 3D awareness, we train our framework on a view-rich human video dataset encompassing diverse viewpoints and camera movements, supplemented with auxiliary decoders that provide early-stage geometric alignment supervision to accelerate spatial understanding (\cref{sec:method_training}).

\subsection{Preliminary}
\label{sec:method_preliminary}
\myparagraph{Video Generation Backbone.} Our video generation model adopts the latent diffusion model (LDM) paradigm, utilizing a causal 3D VAE for video compression into latent space and a generative backbone for latent sequence modeling. 
The backbone is a DiT-based architecture~\cite{sd3} pretrained on text-to-video and image-to-video tasks, comprising multiple DiT blocks that interleave full self-attention and Feed-Forward Networks (FFN).
The reference image is incorporated by concatenating its latent tokens with noised video tokens, facilitating cross-modal interaction among reference, video, and text tokens through the full self-attention.
During training, we adopt a flow-based diffusion process~\cite{liu2022flow,lipman2022flow} that progressively adds noise to the target video latents, and the model is optimized using a v-prediction objective.

\myparagraph{Parametric 3D Human Model.} 
SMPL~\cite{SMPL:2015} represents a full-body human mesh using a compact set of parameters, including shape coefficients $\bm{\beta}_b$ and pose parameters $\bm{\theta}_b$ that describe the articulated body configuration.
Similarly, MANO~\cite{MANO:SIGGRAPHASIA:2017} models the articulated hand using an analogous formulation, with hand-shape parameters $\bm{\beta}_h$ and hand-pose parameters $\bm{\theta}_h$.
Despite their well-known limitations in expressiveness and depth ambiguity, these parametric models provide robust 3D geometric priors that we leverage for early-stage auxiliary supervision.

\subsection{End-to-end Framework with Implicit View-Agnostic Motion Control}
\label{sec:method_pipelie}
As illustrated in~\cref{fig:pipeline}, our framework features a motion encoders that extracts motion representations $\bm{z}$ from the driving video $\bm{V}_D$, which condition a pretrained DiT-based video generator.
The reference image $\bm{I}_R$ and accompanying text prompt $T$ are also provided as token sequences to the video generator, which produces an output video depicting the reference subject in $\bm{I}_R$ reenacting the motion from $\bm{V}_D$ under camera trajectories guided by $T$.
Unlike previous methods that rely on external pose estimation, our framework jointly optimizes the motion encoder and the video generator in an end-to-end manner, learning semantically rich motion representations naturally aligned with the generator's inherent spatial and motion priors.

\myparagraph{Implicit Motion Encoder.}
The core insight of our motion encoder design is to encourage view-agnostic representation learning that captures the semantically rich dynamics of 3D human motion, going beyond the superficial patterns observable in 2D projections.
Following~\cite{yu2024image}, we design our motion encoder as a Transformer-based 1D tokenizer.
Each driving frame is patchified into visual tokens and concatenated with $K(=5$ in our work) learnable latent tokens, which interact through several attention layers.
Only the output latent tokens are retained as the motion representation.
By compressing into compact 1D motion tokens, we enforce a semantic bottleneck that eliminates 2D structural information, including both appearance details and view-specific pose configurations, while focusing on the intrinsic semantics of spatial motion.

To encourage view-agnostic motion representation learning, we apply random perspective transformations to the driving frames before motion encoding to introduce motion-invariant augmentations, 
which to some extent decouples the spatial motion from its view-specific 2D projection. 
Additionally, similar to~\cite{zhao2025xnemo,song2025xuni}, we employ appearance augmentations (\eg, color jittering and lightweight spatial transforms) to prevent identity leakage from the driving frames.
Through this design, the motion encoder is encouraged to focus purely on the intrinsic dynamics of 3D spatial motion while avoiding both appearance leakage and view-specific pose constraints.

\myparagraph{View-Agnostic Cross Attention Conditioning.}
Instead of converting our motion representations into view-dependent 2D spatially-aligned control using explicit camera parameters as in most existing works, we simply employ cross-attention to inject motion representations directly into the generator.
This achieves flexible semantic-level interaction between visual and motion modalities without rigid spatial constraints.
To be specific, we append a cross-attention layer after each full self-attention in the DiT generator, where only video tokens attend to the motion tokens, while text tokens remain unchanged.

\myparagraph{Text-Guided Camera Control.}
Our view-agnostic motion representations and semantic-level conditioning naturally coexist with the generator’s native text-driven camera control capability.
Consequently, beyond motion control, our framework readily supports flexible viewpoint manipulation by simply augmenting the text prompt $T$ with camera-movement descriptions, which interact with visual tokens through the same native mechanism as in the original DiT-based generator.

\myparagraph{Dual-Scale Motion Encoding.}
Considering that a single compact representation struggles to capture both global body movements and fine-grained hand dynamics, we follow~\cite{song2025xuni} and employ two motion encoders: a body encoder $\mathcal{E}_b$ for coarse body motion and a hand encoder $\mathcal{E}_h$ for detailed gestures.
The resulting motion tokens are concatenated and jointly injected into the generator via cross-attention, enabling unified motion control over both full-body articulation and fine-grained hand movements.

\begin{figure}[t]
  \centering
   \includegraphics[width=\linewidth]{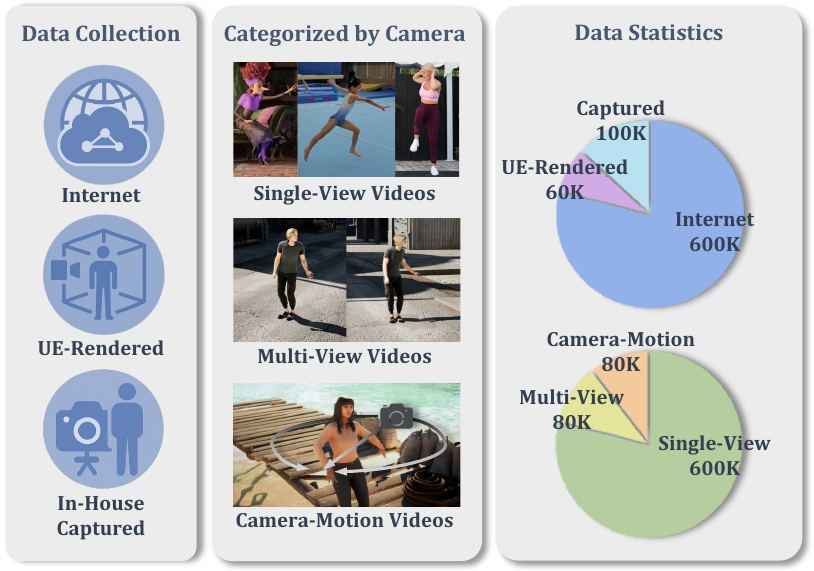}
   \caption{
   \textbf{Our collected view-rich dataset} combines internet videos, UE renderings, and in-house captures, covering camera categories including single-view, multi-view, and camera-motion sequences.
   High-quality large-scale single-view footage exposes the model to diverse human motions, while complementary multi-view data provides consistent cross-view observations that are crucial for learning genuine 3D-aware motion representations.
   }
   \label{fig:dataset}
\end{figure}

\subsection{3D-Aware Training with View-Rich Data}
\label{sec:method_training}
While our motion encoder with distilled 1D tokens effectively filters redundant 2D spatial structures and captures semantically rich motion dynamics for expressive motion generation, this design alone does not guarantee genuine 3D motion understanding.
When trained solely through same-view reconstruction, the model can achieve satisfactory results by merely learning view-dependent 2D motion patterns, as reproducing motion under identical viewpoints requires no true spatial reasoning.
This reveals a fundamental limitation: \emph{without more challenging supervision, the model lacks incentive to develop 3D awareness beyond the 2D projection domain.}
To overcome this issue, we introduce view-rich data supervision that imposes a more demanding learning objective, compelling the model to reason about motion as it truly occurs in 3D space, invariant to viewpoint changes rather than as isolated 2D observations.

\myparagraph{View-Rich Dataset Construction.}
To enable comprehensive supervision for learning both expressive and 3D-aware motion representations, we construct a large-scale dataset with diverse camera configurations.
From the perspective of supervision objectives, our data serves three distinct purposes as shown in~\cref{fig:dataset}: 1) same-view reconstruction, where each motion-viewpoint pair is unique, enabling self-supervised learning of expressive motion dynamics; 2) multi-view motion reproduction, utilizing synchronized captures from fixed camera arrays for identical motions to enforce consist 3D motion learning across viewpoints; and 3) motion reproduction under moving cameras, featuring identical motions captured with different camera trajectories to decouple motion from viewpoint changes and support text-guided camera control.

To balance realism, diversity, and 3D consistency, we integrate three complementary data sources: 1) large-scale internet videos that provide diverse human motions for learning realistic dynamics, though limited to single-view supervision; 2) synthetic sequences rendered with Unreal Engine 5 (UE5) from~\cite{bai2025recammaster,wang2025cinemaster,luo2025camclonemaster}, which offer precise motion captures under varied camera trajectories, despite potential domain gaps from real-world videos; 3) real-world multi-view captures, including both open-source datasets and our proprietary recordings, combining fixed multi-camera arrays and dynamic camera trajectories, which provide authentic 3D supervision within the real-world video domain.
The detailed composition of our dataset is illustrated in~\cref{fig:dataset}.
In practice, internet videos dominate in scale and drive the learning of natural, expressive motion patterns, whereas the multi-view and camera-trajectory data, though smaller in quantity, play a crucial role in fostering genuine 3D spatial understanding.

To obtain text descriptions for camera viewpoints and movements, we employ Qwen2.5-VL~\cite{bai2025qwen2} to both annotate internet videos  and convert predefined camera configurations from synthetic and real-world captured data into unified text prompts.

\myparagraph{Training Strategies.}
With the constructed dataset, we enable 3D-aware training under view-rich supervision that includes both reconstruction and cross-view motion reproduction objectives.
Specifically, given a driving video $\bm{V}_D$, we supervise the model output with either $\bm{V}_D$ itself (reconstruction) or corresponding videos $\bm{V}'_D$ of the same motion captured from a different viewpoint or camera trajectory (cross-view reproduction).
The reference image is taken as the first frame of the supervision target, which automatically aligns the generated motion with the reference subject's facing direction---eliminating the need for explicit SMPL-to-image alignment or camera regression as in~\cite{cao2025uni3c}.

In practice, we adopt a progressive multi-stage training strategy with varying data mixtures based on supervision objectives.
In the first stage, we exclusively use single-view data for self-reconstruction, exposing the model to diverse and expressive motion dynamics and enabling stable initialization of implicit motion learning.
The second stage introduces a balanced mixture of reconstruction and cross-view motion reproduction, gradually transitioning the learned representations from 2D dynamics toward 3D spatial semantics.
Finally, the third stage focuses entirely on multi-view and camera-motion data to strengthen the view-agnostic nature of the learned motion features and enhance compatibility with flexible text-guided camera control.

\myparagraph{Geometric Supervision with Auxiliary Decoders.}
In our early experiments, we observed that direct end-to-end training often leads to slow and unstable convergence, especially after introducing cross-view supervision.
This partly arises because the diffusion loss distributes uniformly across pixels, lacking targeted emphasis on motion-specific semantics.
Moreover, the powerful DiT backbone tends to exploit its inherent motion priors to generate plausible videos from single images during early training, thereby reducing reliance on the encoded motion representations, which consequently receive weak gradient feedback.

To address this challenge, we introduce auxiliary geometric supervision to facilitate motion representation learning.
Specifically, we employ a lightweight MLP-based geometric decoder $\mathcal{D}_g$ that processes the concatenated motion representations $\mathbf{z}=[\mathbf{z}_b;\mathbf{z}_h]$ to predict pose parameters $\bm{\theta}=[\bm{\theta}_b; \bm{\theta}_h]$, using pseudo ground-truth annotations derived from off-the-shelf SMPL and MANO estimators~\cite{patel2025camerahmr,potamias2025wilor}.
Notably, we exclude the global root orientation during supervision to ensure view-agnostic learning.

Despite limitations in depth ambiguity and expressiveness, this parametric 3D geometric supervision effectively transfers robust spatial motion priors through the lightweight, easily optimized auxiliary decoder, providing a well-initialized motion distribution for subsequent learning.
In practice, we apply auxiliary supervision during the first stage and the early part of the second stage, with its loss weight annealed progressively as training proceeds. The supervision is then completely removed for the remaining steps of the second stage and the entirety of the third stage.
This schedule allows the model to evolve from geometry-guided initialization to learning motion representations that align with the DiT's perceptual and generative capabilities, ultimately achieving superior 3D-aware motion understanding supported by view-rich supervision.

\section{Experiments}
\subsection{Experimental Setups}
\myparagraph{Implementation Details.}
We train our \methodname{} on the dataset described in~\cref{sec:method_training}, using 121-frame video clips resized to a target area of $480\times854$ pixels while preserving original aspect ratios. Training is performed with a total batch size of 64 using the Adam optimizer with a learning rate of 1e-5. The three training stages run for 10K, 15K, and 5K steps, completing in approximately three days. 

\myparagraph{Evaluation Data and Metrics.} 
We evaluate our method on 50 videos from TikTok dataset~\cite{jafarian2021tiktok} and 100 videos collected from the internet.
Following~\cite{hu2024animate}, we use PSNR, SSIM, LPIPS, and FID to measure per-frame visual quality, and adopt FVD to evaluate the overall video fidelity.

\subsection{Quantitative Evaluation}
\label{sec:quantitative_eval}
\begin{table*}[t]
\centering
\caption{Quantitative evaluation and user study results of MOS with 95\% confidence intervals. Top two are noted as \first{first}, \second{second}.}
\label{tab:comp}
\resizebox{\textwidth}{!}{
\begin{tabular}{@{}lccccccccc@{}}
\toprule
\multicolumn{1}{c}{} & \multicolumn{5}{c}{Quantitative Evaluation} & \multicolumn{4}{c}{User Study} \\ \cmidrule(lr){2-6} \cmidrule(lr){7-10}
\multicolumn{1}{c}{\multirow{-2}{*}{Method}} & SSIM $\uparrow$ & PSNR $\uparrow$ & LPIPS $\downarrow$ & FID $\downarrow$ & FVD $\downarrow$ & Accuracy $\uparrow$ & Naturalness $\uparrow$ & 3D Plausibility $\uparrow$ & Overall $\uparrow$ \\ \midrule
AnimateAnyone & 0.7325 & 17.21 & 0.2754 & 68.72 & 862.5 & \cellcolor[HTML]{FFDA8F}4.13$\pm$0.12 & 4.00$\pm$0.09 & 3.20$\pm$0.14 & 3.76$\pm$0.10 \\
MimicMotion & 0.7051 & 16.83 & 0.3286 & 62.45 & 628.2 & 3.84$\pm$0.14 & \cellcolor[HTML]{FFDA8F}4.15$\pm$0.06 & 3.01$\pm$0.12 & 3.84$\pm$0.10 \\
MTVCrafter & \cellcolor[HTML]{FE996B}0.7489 & \cellcolor[HTML]{FE996B}18.03 & \cellcolor[HTML]{FFDA8F}0.2542 & 57.21 & 379.6 & 3.69$\pm$0.09 & 3.98$\pm$0.10 & 3.69$\pm$0.12 & 4.19$\pm$0.09 \\
Uni3C & 0.7185 & 17.53 & 0.2639 & \cellcolor[HTML]{FFDA8F}41.28 & \cellcolor[HTML]{FFDA8F}321.9 & 3.72$\pm$0.06 & 4.03$\pm$0.06 & \cellcolor[HTML]{FFDA8F}3.97$\pm$0.10 & \cellcolor[HTML]{FFDA8F}4.24$\pm$0.06 \\
\textbf{Ours} & \cellcolor[HTML]{FFDA8F}0.7390 & \cellcolor[HTML]{FFDA8F}17.96 & \cellcolor[HTML]{FE996B}0.2206 & \cellcolor[HTML]{FE996B}36.92 & \cellcolor[HTML]{FE996B}297.4 & \cellcolor[HTML]{FE996B}4.28$\pm$0.08 & \cellcolor[HTML]{FE996B}4.18$\pm$0.06 & \cellcolor[HTML]{FE996B}4.05$\pm$0.09 & \cellcolor[HTML]{FE996B}4.38$\pm$0.08 \\ \bottomrule
\end{tabular}
}
\end{table*}

We compare \methodname{} with state-of-the-art human image animation methods, including both 2D pose-based approaches (AnimateAnyone~\cite{hu2024animate}, MimicMotion~\cite{zhang2024mimicmotion}) and 3D SMPL-based methods (Uni3C~\cite{cao2025uni3c}, MTVCrafter~\cite{ding2025mtvcrafter}), to validate the effectiveness of our implicit 3D-aware motion modeling and reenactment.
Considering that most existing baselines do not support camera manipulation, we evaluate 3DiMo under prompts specifying a \emph{static camera}.

As shown in~\cref{tab:comp} (Col. 2-6), our method surpasses all baselines on LPIPS, FID, and FVD, indicating superior visual quality and motion control. Although our SSIM and PSNR are slightly lower than MTVCrafter, this is expected: these pixel-wise metrics are sensitive to minor viewpoint deviations while some evaluation videos contain weak, unintended camera motions. Competing methods mainly operate within a 2D alignment paradigm and simply reproduce these motions, while our text-driven \emph{static camera} prompt suppresses such drift to maintain geometric consistency. This yields perceptually better results but introduces small pixel-wise discrepancies from the ground truth.

\subsection{Qualitative Evaluation}
\begin{figure}[t]
  \centering
   \includegraphics[width=\linewidth]{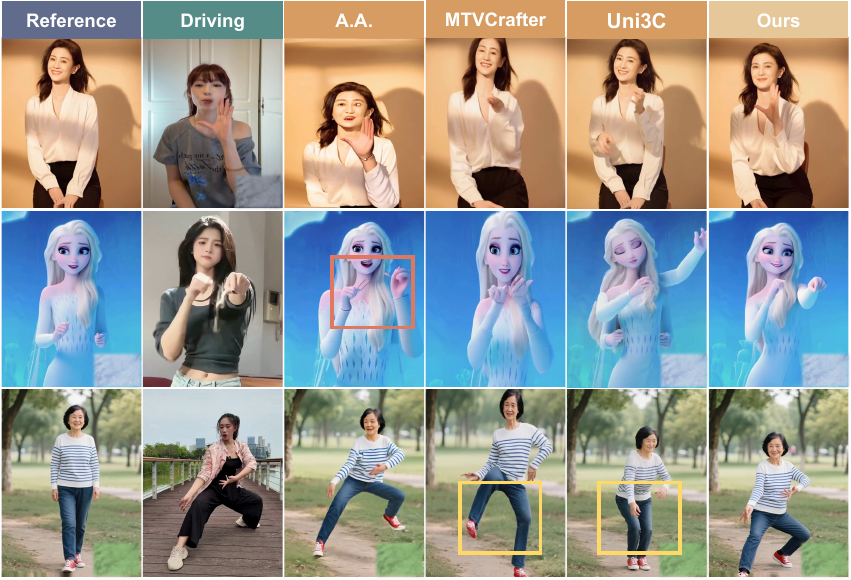}
   \caption{
\textbf{Visualization comparisons with baselines.} 
Red and yellow bounding boxes highlight depth ambiguities and inaccurate poses, respectively. 
``A.A.'' denotes AnimateAnyone. 
Our method produces accurate and 3D-plausible motion reenactment videos.
}
   \label{fig:comp}
\end{figure}

\myparagraph{Comparison with SOTAs.}
\cref{fig:comp} presents the quantitative comparisons, where we configure our method with a static camera as described in~\cref{sec:quantitative_eval}.
Our implicit 3D-aware approach achieves precise motion control and expressive dynamics, producing high-fidelity and physically plausible human videos.
In contrast, 2D pose-based methods often yield incorrect limb depth ordering due to their lack of geometric awareness, while SMPL-based approaches struggle to maintain accurate pose estimation and control under complex motions.
More visualization results and extended comparisons are provided in the supplementary material.

\myparagraph{More Results of View-Adaptive Motion Control.}
To further demonstrate our model's capability in modeling 3D motion, Figure~\cref{fig:teaser} showcases several motion control results under diverse text-guided camera configurations.
Our approach inherits the DiT backbone's native ability for text-driven camera manipulation, while simultaneously achieving precise video-driven motion control that preserves physical plausibility and spatial consistency across dynamic camera trajectories and varying viewpoints.

\myparagraph{User Study.}
We further conduct a user study involving 30 participants, where each participant evaluates 10 cross-identity animation videos generated by each method.
We collect mean opinion scores (MOS) based on a 5-point Likert scale across four aspects: motion accuracy, motion naturalness, 3D physical plausibility, and overall visual quality.
The results, summarized in~\cref{tab:comp} (Col. 7-10), show that our method consistently outperforms all existing baselines, especially in motion naturalness and physical plausibility, which emphasize spatial relationships and realistic dynamics.
These strongly support that our learned motion representation, aligned with the large-scale pretrained video generator's spatial and motion priors, achieves more expressive and 3D-aware motion modeling and control compared to approaches that rely on externally predefined parameters.

\subsection{Ablation Study and Analysis}
\begin{figure}[t]
  \centering
   \includegraphics[width=\linewidth]{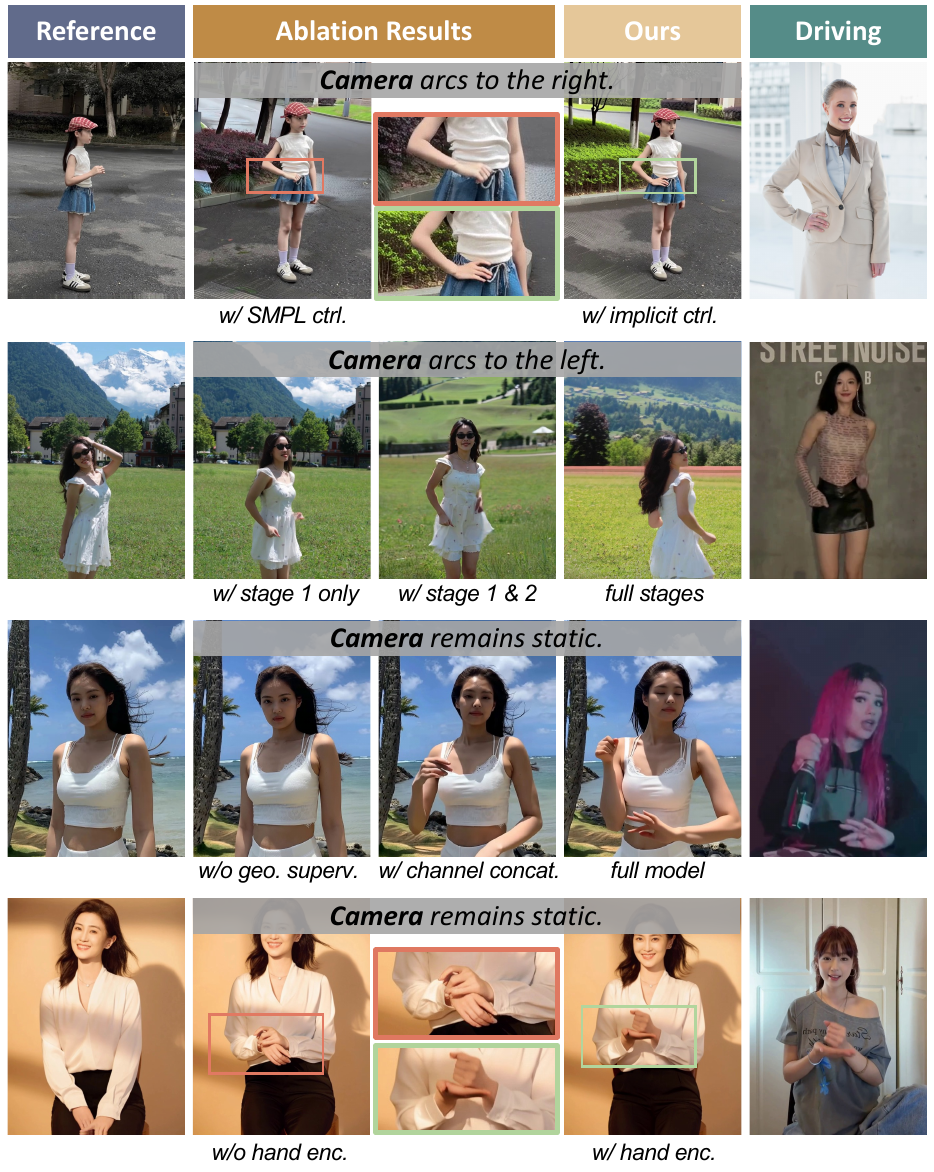}
   \caption{
\textbf{Visualizations of ablation results.}
Using SMPL poses as motion representation introduces typical depth ambiguity errors.
Removing any view-rich data supervision impairs camera control.
Removing auxiliary geometric supervision or using channel concatenation causes training instability and quality degradation.
Without the hand encoder, fine-grained hand motions are lost.
}
   \label{fig:ablation}
\end{figure}

\begin{table}[t]
\centering
\caption{Ablation results. Top two are noted as \first{first}, \second{second}.}
\label{tab:ablation}
\resizebox{.5\textwidth}{!}{
\begin{tabular}{@{}lccccc@{}}
\toprule
\multicolumn{1}{c}{Method} & SSIM $\uparrow$ & PSNR $\uparrow$ & LPIPS $\downarrow$ & FID $\downarrow$ & FVD $\downarrow$ \\ \midrule
w/ SMPL ctrl. & 0.724 & 17.1 & 0.238 & 39.7 & 348.2 \\
w/ stage 1 only & \cellcolor[HTML]{FE996B}0.745 & \cellcolor[HTML]{FE996B}18.3 & \cellcolor[HTML]{FE996B}0.220 & 40.5 & 305.4 \\
w/ stage 1 \& 2 & 0.723 & 17.9 & 0.221 & 38.2 & 314.5 \\
w/ channel concat. & 0.703 & 16.8 & 0.304 & 48.2 & 395.6 \\
w/o geo. superv. & 0.684 & 15.8 & 0.347 & 51.3 & 383.1 \\
w/o  hand enc. & 0.726 & 17.5 & 0.234 & \cellcolor[HTML]{FFDA8F}38.1 & \cellcolor[HTML]{FFDA8F}298.7 \\
\textbf{Full Model} & \cellcolor[HTML]{FFDA8F}0.739 & \cellcolor[HTML]{FFDA8F}18.0 & \cellcolor[HTML]{FFDA8F}0.221 & \cellcolor[HTML]{FE996B}36.9 & \cellcolor[HTML]{FE996B}297.4 \\ \bottomrule
\end{tabular}
}
\end{table}

\myparagraph{SMPL \vs Our Implicit Motion Representations.} 
The core of our approach lies in the end-to-end learning of an implicit 3D-aware motion representation.
For comparison, we also consider a variant that directly uses SMPL pose coefficients $\theta_{\text{body}}$ as the motion representation, mapped through an MLP to match the token dimensionality before being injected into the generator via the same cross-attention mechanism.

As shown in~\cref{fig:ablation}, for a frontal one-hand-on-hip driving motion, the SMPL-based variant fails to maintain correct hand-hip contact from the side view.
In contrast, our learned motion representation correctly preserves this physical relationship, effectively resolving the depth ambiguity that commonly occurs in parametric reconstructions.
This demonstrates that our motion representation, distilled from the pretrained video generator, exhibits superior 3D spatial awareness aligned with real-world priors compared to off-the-shelf parametric reconstruction.

\myparagraph{Multi-Stage View-Rich Learning.}
We further analyze the effect of our multi-stage training strategy under a leftward arcing camera trajectory.
In the first stage with single-view reconstruction, the model learns diverse motion patterns but tends to collapse into 2D projections, often failing to follow text-guided camera motions.
Introducing view-rich data in the second stage helps decouple motion and viewpoint, establishing initial 3D awareness; however, the camera motion sometimes only affects the background while the subject remains front-facing, indicating incomplete semantic interaction between motion and camera control.
The third-stage refinement, trained solely on view-rich data, further strengthens this interaction, enabling the model to accurately follow camera trajectories while maintaining spatially consistent human motion and background rendering.

\myparagraph{Conditioning Mechanism.}
We replace cross-attention with channel concatenation as an alternative conditioning approach.
This substitution shows significantly degraded motion control capability, demonstrating that cross-attention is more suitable for semantic-rich interaction between our motion representations and the generator.

\myparagraph{Auxiliary Geometric Supervision.}
Removing the auxiliary geometric supervision during early training leads to unstable convergence and collapsed motion control, demonstrating that this supervision provides crucial initialization for learning meaningful motion representations.

\myparagraph{Dual-scale motion encoders.}
Omitting the hand motion encoder results in loss of fine-grained hand control, confirming its necessity for complete motion representation.

\cref{tab:ablation} further shows that removing most components consistently degrades visual quality, confirming their necessity for producing high-fidelity human motion videos. 
Discarding the last two view-rich training stages yields a slight but negligible improvement in visual metrics; however, these stages are essential for endowing the model with genuine 3D-aware motion understanding—precisely the core capability required to address our proposed 3D-aware motion control problem.

\section{Conclusion}
In this work, we present \methodname{}, an end-to-end framework for 3D-aware human motion control that learns to align with a video generator’s intrinsic spatial priors rather than relying on explicit 3D parametric reconstruction. By jointly training a motion encoder with the pretrained video generation model and designing it to discard view-dependent layouts, \methodname{} aligns the two models and establishes the capacity for view-agnostic motion representation. When further trained under view-rich data supervision, the framework internalizes 3D spatial motion understanding directly from 2D observations. Combined with lightweight geometric initialization that is gradually annealed away, \methodname{} ultimately develops robust and expressive 3D-aware motion representations without relying on external estimates.
Experiments demonstrate that \methodname{} faithfully reproduces driving motions under flexible text-driven camera control and consistently outperforms both 2D- and 3D-based baselines in motion fidelity and visual quality.

{
    \small
    \bibliographystyle{ieeenat_fullname}
    \bibliography{main}
}

\clearpage
\setcounter{page}{1}

\setcounter{section}{0}   %
\setcounter{figure}{0}    %
\setcounter{table}{0}     %
\setcounter{equation}{0}  %

\renewcommand\thesection{\Alph{section}}
\renewcommand\thefigure{S\arabic{figure}}  %
\renewcommand\thetable{S\arabic{table}}    %
\renewcommand\theequation{S\arabic{equation}} %

\twocolumn[{
    \begin{center}
        \textbf{\Large 3D-Aware Implicit Motion Control for View-Adaptive Human Video Generation} \\[0.5em]
        \textbf{\large Supplementary Material}
    \end{center}
}]

\begin{figure*}[t]
\centering
\includegraphics[width=\textwidth]{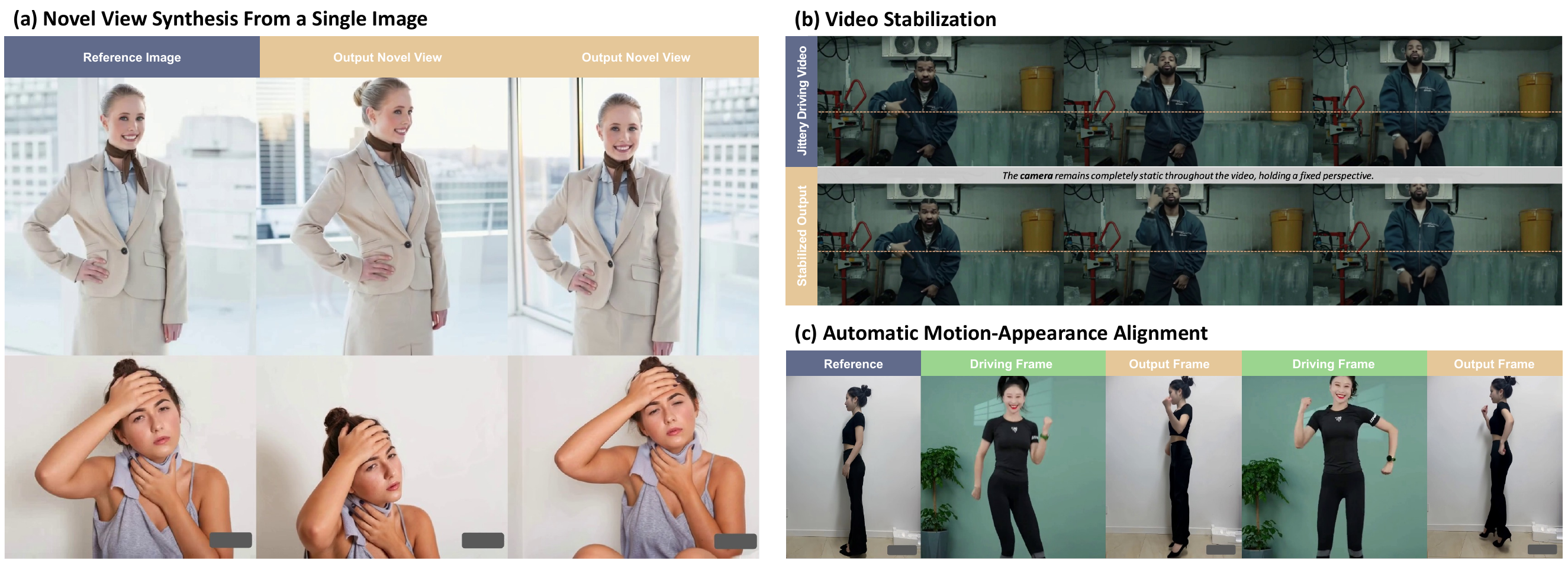}
\caption{
\textbf{Broader applications of \methodname{}. } We demonstrate the versatility of our framework on three downstream tasks: (a) single-image novel view synthesis by enforcing static motion; (b) video stabilization by suppressing camera jitter from the driving video; and (c) automatic motion-appearance alignment without explicit calibration.
}
\label{fig:application}
\end{figure*}

In this supplementary document, we provide additional details to support the main paper, organized as follows:
\begin{itemize}
    \item \textbf{Section~\ref{sec:ethics}:} Ethical considerations regarding human video generation and our commitment to responsible research.
    \item \textbf{Section~\ref{sec:applications}:} Demonstrations of broader applications, including single-image novel view synthesis, video stabilization, and automatic motion alignment.
    \item \textbf{Section~\ref{sec:data_details}:} Detailed specifications of our in-house data collection pipeline and the definition of camera trajectories.
    \item \textbf{Section~\ref{sec:limitations}:} A discussion on current limitations and potential directions for future work.
    \item \textbf{Section~\ref{sec:implementation}:} Technical implementation details.
\end{itemize}

\section{Ethical Considerations}
\label{sec:ethics}
The rapid progress in AI-driven human video generation offers significant potential for digital entertainment, virtual reality, and human-centric research. However, the ability to synthesize highly realistic human content also brings forth important ethical considerations, such as the potential for privacy violations, intellectual property concerns, and the risk of creating deceptive ``deepfake'' media.

As with many generative technologies, there is a possibility that these methods could be misused to create content without the consent of the individuals involved. Addressing these risks requires the collective development of ethical guidelines and legal frameworks. In this work, we are committed to responsible research practices. All processed data, models, and results are intended strictly for academic purposes and are not authorized for commercial use or the creation of harmful content. We believe that by adhering to these principles, the proper application of such techniques will continue to positively enhance research in computer graphics and artificial intelligence.

\section{Broader Applications}
\label{sec:applications}
Benefiting from the implicit 3D motion reasoning and flexible text-driven camera control of \methodname{}, our approach generalizes effectively to specific downstream tasks, as shown in Fig.~\ref{fig:application}.

\myparagraph{Human Novel View Synthesis From a Single Image.}
While traditional novel view synthesis (NVS) typically requires reconstructing 3D scenes from reference images to render new angles, \methodname{} achieves human-specific single-image NVS through a straightforward inference strategy. We construct a driving video by repeating the reference frame (implying zero motion) and pair it with text prompts describing camera trajectories (e.g., ``\textit{camera rotates in a circular path around the woman}'').
Although pretrained I2V foundation models theoretically support this via prompts specifying camera movement alongside ``\textit{static subject},'' they suffer from significant limitations in practice. As noted by~\cite{wiedemer2025video}, these models tend to hallucinate motion, failing to keep the subject strictly stationary. Moreover, we observe that base I2V models often confuse camera control with background animation rather than performing true geometric view synthesis. By leveraging our view-agnostic motion representation and the model's improved 3D awareness, \methodname{} overcomes these ambiguities to produce consistent novel-view generations.

\myparagraph{Video Stabilization.}
Capturing stable footage during dynamic recording is often challenging. Video stabilization aims to smooth out camera jitters to obtain high-quality, steady sequences. In human-centric scenarios, \methodname{} effectively performs this task. By utilizing the first frame of the shaky video as the reference image and the full video as the driving signal, we can feed the model a prompt such as ``\textit{camera remains static}.'' This instructs the generator to reconstruct the underlying human motion from a fixed viewpoint, effectively eliminating the original camera shake while preserving the subject's dynamics.

\myparagraph{Automatic Motion-Image Alignment.}
Conventional motion transfer methods, particularly 2D-based approaches, rigidly impose the absolute orientation of the driving video onto the reference subject. This often leads to unnatural transitions when the driving and reference subjects have different initial facing directions (e.g., a side-view driver controlling a front-view reference). 
In contrast, because \methodname{} extracts a view-agnostic implicit motion representation, it naturally aligns the driving motion with the reference subject's initial orientation. Our model transfers the relative 3D dynamics rather than the absolute 2D projection, eliminating the need for manual camera calibration or explicit root-rotation alignment required by SMPL-based methods.

\section{In-House Data Acquisition Setup}
\label{sec:data_details}
Our in-house data capture involves a three-camera array positioned at diverse angles relative to the subject. For every captured performance, each camera is assigned a camera motion type sampled randomly from the following categories:
\begin{itemize}
    \item \textbf{Static Variants:} Static, Handheld Static.
    \item \textbf{Linear Translations:} Move Forward, Move Back, Move Left, Move Right, Move Up, Move Down.
    \item \textbf{Zoom Actions:} Zoom In, Zoom Out, Rapid Zoom In, Rapid Zoom Out, Handheld Zoom In, Aerial Pull-out.
    \item \textbf{Complex Trajectories:} Vertigo In, Vertigo Out, Dynamic Zoom Swing, Arc Left (variable angles, e.g., $30^\circ, 45^\circ$), and Arc Right (variable angles, e.g., $30^\circ, 45^\circ$).
\end{itemize}
By pairing identical human motions with diverse, non-correlated camera trajectories across three views, we maximize the supervision signal for view-agnostic motion learning.

\section{Limitations and Future Work}
\label{sec:limitations}
Despite the significant advancements \methodname{} achieves in view-adaptive human video generation, several limitations remain to be addressed in future research.

\myparagraph{Resolution and Fine-Grained Details.}
Currently, our framework operates at a resolution of 480p. While this is sufficient for capturing global motion dynamics, it imposes a bottleneck on high-frequency details. Specifically, in full-body shots where the subject occupies a relatively small proportion of the frame, the limited pixel budget can lead to artifacts, such as blurred facial features or a lack of texture in hand details. 
Future iterations could address this by scaling up the framework to higher-resolution DiT backbones (e.g., 720p or 1080p) or incorporating cascaded super-resolution modules to enhance local details in small-scale regions.

\myparagraph{Complex Human-Object Interactions.}
Since our motion encoders are explicitly designed to distill human body and hand dynamics, the current framework does not explicitly model the motion of external objects or props (e.g., a person holding a bag or riding a bicycle). Consequently, while the human motion is faithfully reproduced, the interaction with held objects may sometimes be hallucinated. 
Extending the implicit motion encoding mechanism to handle general dynamic objects or human-scene interactions represents a promising direction for future work.

\section{Implementation Details}
\label{sec:implementation}
We train our \methodname{} using 121-frame video clips resized to a target area of $480\times854$ pixels while preserving original aspect ratios. Training is performed with a total batch size of 64 using the Adam optimizer with a learning rate of 1e-5. The three training stages run for 10K, 15K, and 5K steps, completing in approximately three days. During training, the weight of the auxiliary geometric supervision is linearly annealed from 0.1 to 0 over the first 12K steps.

\end{document}